# Average word length dynamics as indicator of cultural changes in society

## V.V. Bochkarev[1], A.V. Shevlyakova[2] V.D. Solovyev[3]

Kazan Federal University, Kazan, Russia

## Abstract

Dynamics of average length of words in Russian and English is analysed in the article. Words belonging to the diachronic text corpus Google Books Ngram and dated back to the last two centuries are studied. It was found out that average word length slightly increased in the 19th century, and then it was growing rapidly most of the 20th century and started decreasing over the period from the end of the 20th - to the beginning of the 21th century. Words which contributed mostly to increase or decrease of word average length were identified. At that, content words and functional words are analysed separately. Long content words contribute mostly to word average length of word. As it was shown, these words reflect the main tendencies of social development and thus, are used frequently. Change of frequency of personal pronouns also contributes significantly to change of average word length. The other parameters connected with average length of word were also analysed.

## Introduction

Every language reflects a certain way of world perception, in other words presents language view of the world. All the ideas about the world around which are represented by formal word structures and set expressions form a kind of frame which is more or less shared by native speakers.

Up to the present moment, there were two main approaches aimed at studying language view of the world: 1) studying of individual concepts typical for a certain language, studying linguistic and cultural stereotypes of a given language; 2) studying a language dialect on the whole and its prescientific view of the world.

But creation of big diachronic text corpora and development of mathematical methods for data processing enabled absolutely new approach to studying language view of the world.

The research objective is to present frequency-based approach and apply it for survey of average word length change. Some factors which influence average length of words and reflect dynamics of social development will be described in this paper.

Nowadays language dynamics attracts special attention [1-11, 16]. This article gives special focus to such parameters as word length and frequency. Frequency of language units is the basis of usage-based sociolinguistic models of language evolution [12]. Word length is significantly connected with other typological parameters [21]. Word frequency and length are also fundamental parameters in studying of psychological processes of language acquisition and usage [13, 14].

The main object of sociolinguistic studies is innovation diffusion. Some models were introduced [3-7] which aim is to describe processes of linguistic structure changes. The most distinguished result in this area is S-shaped curve of innovation diffusion [10] and of word frequency effect on its change rate [16] (the less frequently a word is used, the more chances it has to be changed).

[1] vladimir.bochkarev@ksu.ru
[2] anna_ling@mail.ru
[3] maki.solovyev@mail.ru

The models are based on the different postulates about nature of innovations spreading. Usually postulate about independence of different words evolution is accepted [4, 12] which enables to model only one word spreading.

Word form change is performed by the process of word frequency usage: frequency of its previous form decays to zero and it disappears, frequency of the new one increases from zero to a certain number. But the process of word frequency change isn`t connected only with the process of substitution of one form by another one. Word frequency can vary under the influence of sociocultural factors (and in such a way reflects these factors) no matter whether it`s used in a certain language any more or not.

Processes of word frequency change under the influence of sociocultural factors are studied in this article. This refers directly to the paper devoted to quantitative analysis of cultural trends [6]. The subject for study is regularities of word average length variations and factors which cause these changes. Average word length is a cumulative parameter which reflects different processes of word frequency changes.

Average word length was counted in different languages though sometimes the data don't match. As for the English language, it makes 5.1 letters [17, 18], as for the Russian language it makes 5.28 [19]. Nevertheless, accurate quantitative analysis of dynamics of this parameter hasn`t been carried out. Obviously, average word length can not only decrease but increase in course of time, in other words a kind of wavelike process takes place. It`s known that word length of any language changes due to global processes of changing the morphological type of a language. All languages change its morphological type in the course of time (agglutinative, inflective, isolating) [15]. These change of morphological type of a language influence average word length but they happen very slowly. As a rule, it takes thousand years or so to make them visible. Typology and historical linguistics research such kind of changes [8].

As the length of words themselves hasn`t changed radically during the two last centuries, the only reason of word length change at this time can be word frequency change (including neologisms and archaisms) which in its turn is caused by pragmatic (sociocultural) and cognitive factors.

Studying dynamics of average word length at relatively short periods of time (decades or several centuries) became possible after creating of big diachronic text corpora.

## Methods and data

Impressive opportunities in this sphere of study opened after creation of digital library Google Books and means of word frequency calculation – Ngram Viewer [20]. In the survey by Michel et al. [6] it is shown how these data can be applied for analysing of cultural trends.

It should be taken into account that Ngram Viewer doesn`t deal with morphological analysis that`s why data concerning not words but word forms are presented. Thus, "word" is further regarded as 'word form'.

Using the total set of n-grams presented on the site average word length can be counted for each year. (Using for example, MathLab.) Though the library in English comprises text dated back to 1520, great amount of texts for reliable statistic computations have appeared only since 1800. (according to the authors` recommendation [20]).

Average word length can be calculated using the following formula:

$$L = \sum_i p_i l_i , \tag{1}$$

where $l_i$ is the length of $i$-th word and $p_i$ is its relative frequency (usage probability). Let`s consider that a particular word frequency ($k$-th word) varies, at that correlations of other words are invariable. According to normalization requirement $\sum_i p_i = 1$ we derive

$$L = p_k l_k + \sum_{i \neq k} p_i l_i = p_k l_k + \tilde{L}_k \left(1 - p_k\right) \tag{2}$$

where $\tilde{L}_k$ is a average word length without considering $k$-th word. Let`s set that the $k$-th word usage frequency varies over $\Delta p_k$. As it seen from the formula (2), the change of average word length is

$$\Delta L = \Delta p_k \left(l_k - \tilde{L}_k\right) \tag{3}$$

Deriving $\tilde{L}_k$-value from the second formula (2), we get

$$\Delta L = \frac{\Delta p_k}{1 - p_k}\left(l_k - L\right) \tag{4}$$

$\Delta L$-value shows the partial contribute of a $k$-th word to the word average length change. For the majority of words, except several most frequently used ones, $p_k << 1$ is realized

$$\Delta L = \Delta p_k \left(l_k - L\right) \tag{5}$$

Thus, increase of the words used, which length is less than current average word length results in decrease of average word length and vice versa

The value derived from the formula (5) enables evaluating the contribution of this or that word to average word length change. Correspondingly, we can distinguish the words which contribute mostly to the change of average word length .

Two different approaches to frequency calculation should be taken into consideration while comparing the further results and diagrams.

Firstly, only words were distinguished from all the 1-grams contained in the base. 1-gram words are those consist of letters or an apostrophe. The list also contains wrongly imaged words and some onomatopoeia-like words but their contribution is relatively small. There are some ngrams which are not words by their nature (numbers etc.) These non-word signs were excluded.

Secondly, it should be considered that Ngram Viewer shows frequencies normalized to the common number of 1-grams but we normalize them to number of 1-gram words.

Ngram Viewer comprises texts written in 7 languages, we deal only with two languages – English and Russian.

Fig. 1 shows average word length distribution in the English language according to Ngram Viewer data.

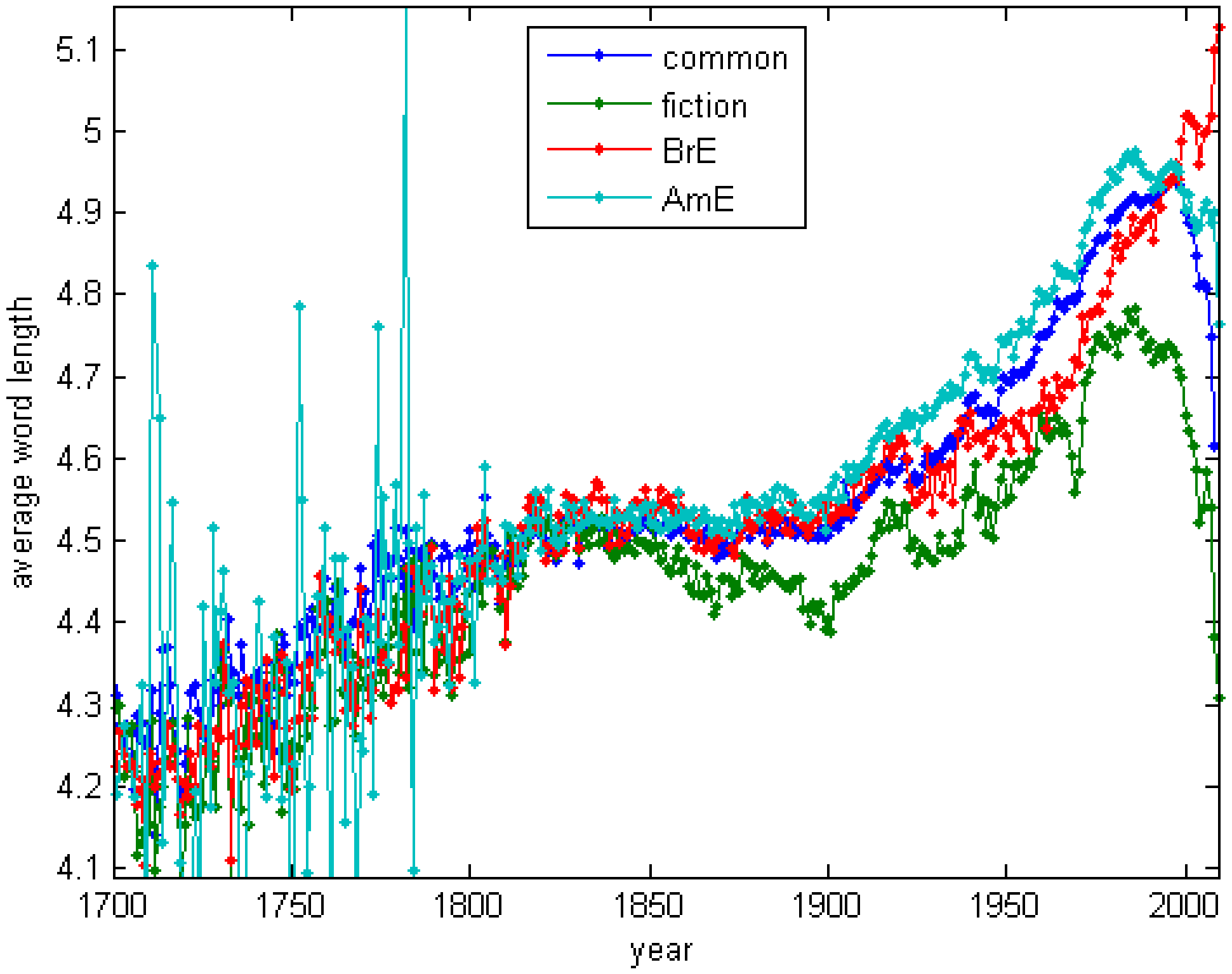

Fig. 1. Average word length in the English language. Different colours indicate the results for the common and fiction bases, and also for British and American bases separately

Three periods can be clearly seen on the graphic: from 1800 till 1900, from 1901 till 1994, from 1995 till 2008. Average word length is almost constant during the 1st period; it increases in linear fashion during the 2nd period and decreases during the 3rd period.

Assume that all the words are divided into two major groups: function words which indicate a grammatical relationship (articles, conjunctions, particles, prepositions etc.) and content words (nouns, verbs, adjectives, numerals). Pronouns and auxiliary verbs were included into the 1st group. Most functional words are short; they consist only of 2-3 letters (less than average word length). Increase in frequency of these word reduces average word length and vice versa. Almost all notional words are long (longer than average word length). That`s why they have opposite influence: increase in frequency of these word lengthens average word length and vice versa.

Frequency variability of functional and notional words is studied separately. Fig. 2 shows changes of average word length of all the examined words and of long and short words separately.

As it can be seen, short words most of which are functional words almost don`t change their length. At the same time the curve shape which shows change of average length of long words most of which are notional tends towards the curve which detects word length of all long words.

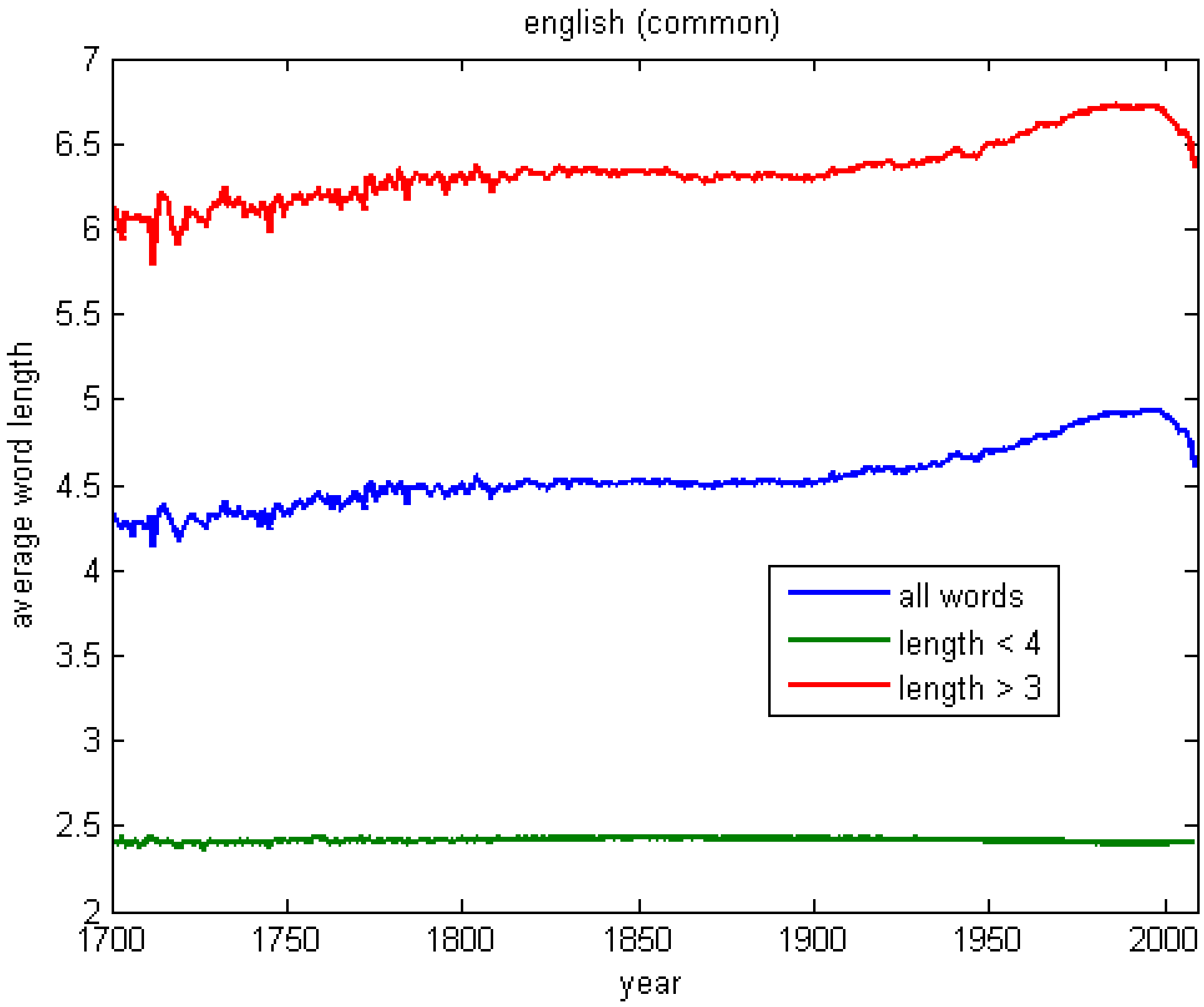

Figure 2. Dynamics of average length of short and long words

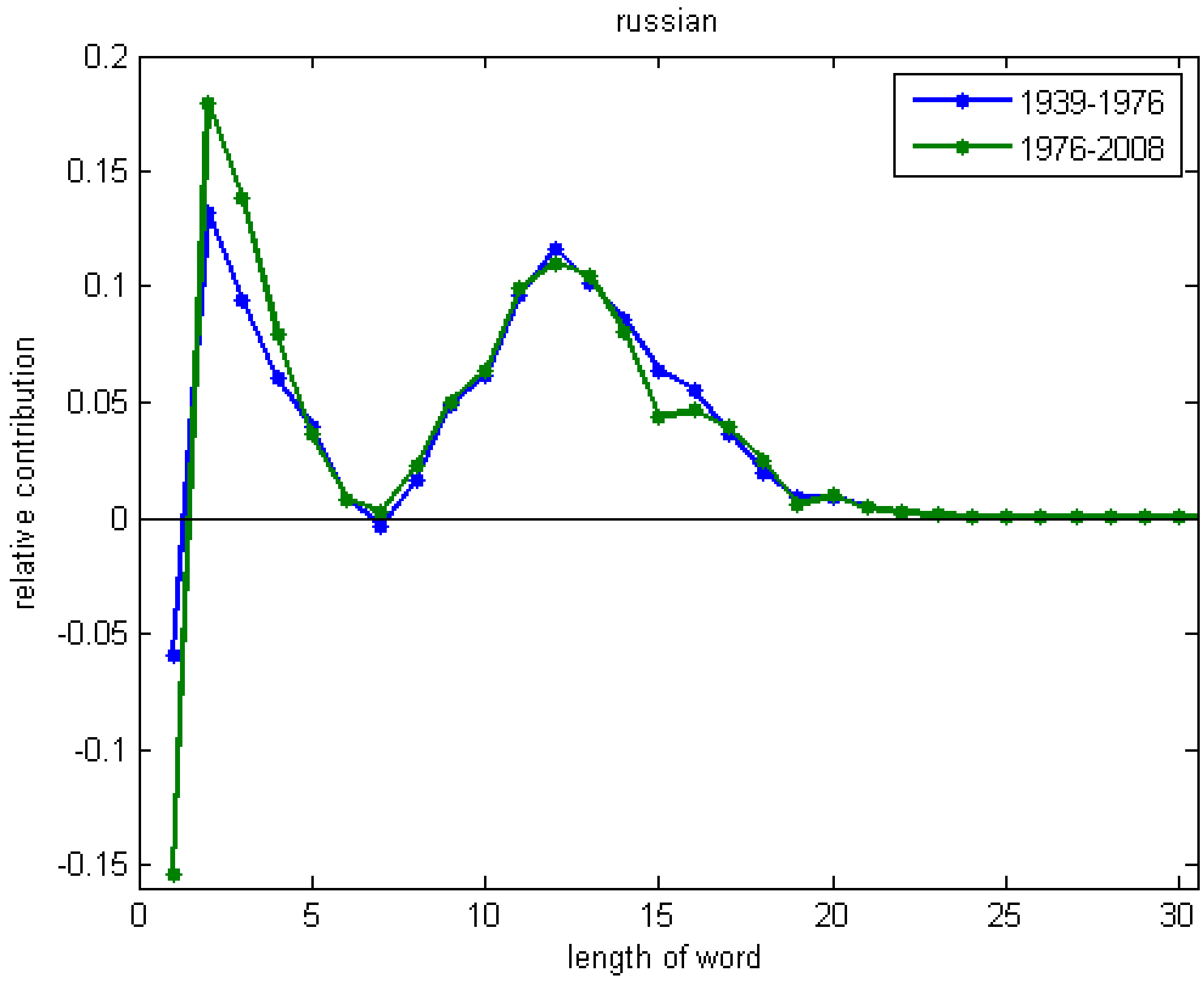

Fig. 3. Relative contribution to average word length change of the words of different length for two time intervals

Fig. 3 shows integral contribution to average word length change of the words of different length in Russian. Peaks sets at two- and twelve-letter words. The period after 1939 is considered to avoid the influence of the orthography reform of 1918. The biggest value of average word length for the Russian language dates back to 1976. Correspondingly, two time periods are distinguished. 1939-1976 when a tendency to average word length increase is observed and 1976-2008 when the opposite tendency takes place.

As the number of long words is greater, their total contribution (approximately 70%) is higher than of short words.

**Hypothesis of social factors influence on notional words**

It can be assumed that increase of average length is determined by vocabulary extension as new words appear. In fact, as there are limited number of short combinations of letters and most of them are used in any language, neologisms such as *computer, the Internet, globalisation* are necessary long. But it`s not so easy as it seems. Fig. 4 shows the amount of words used in different periods of time.

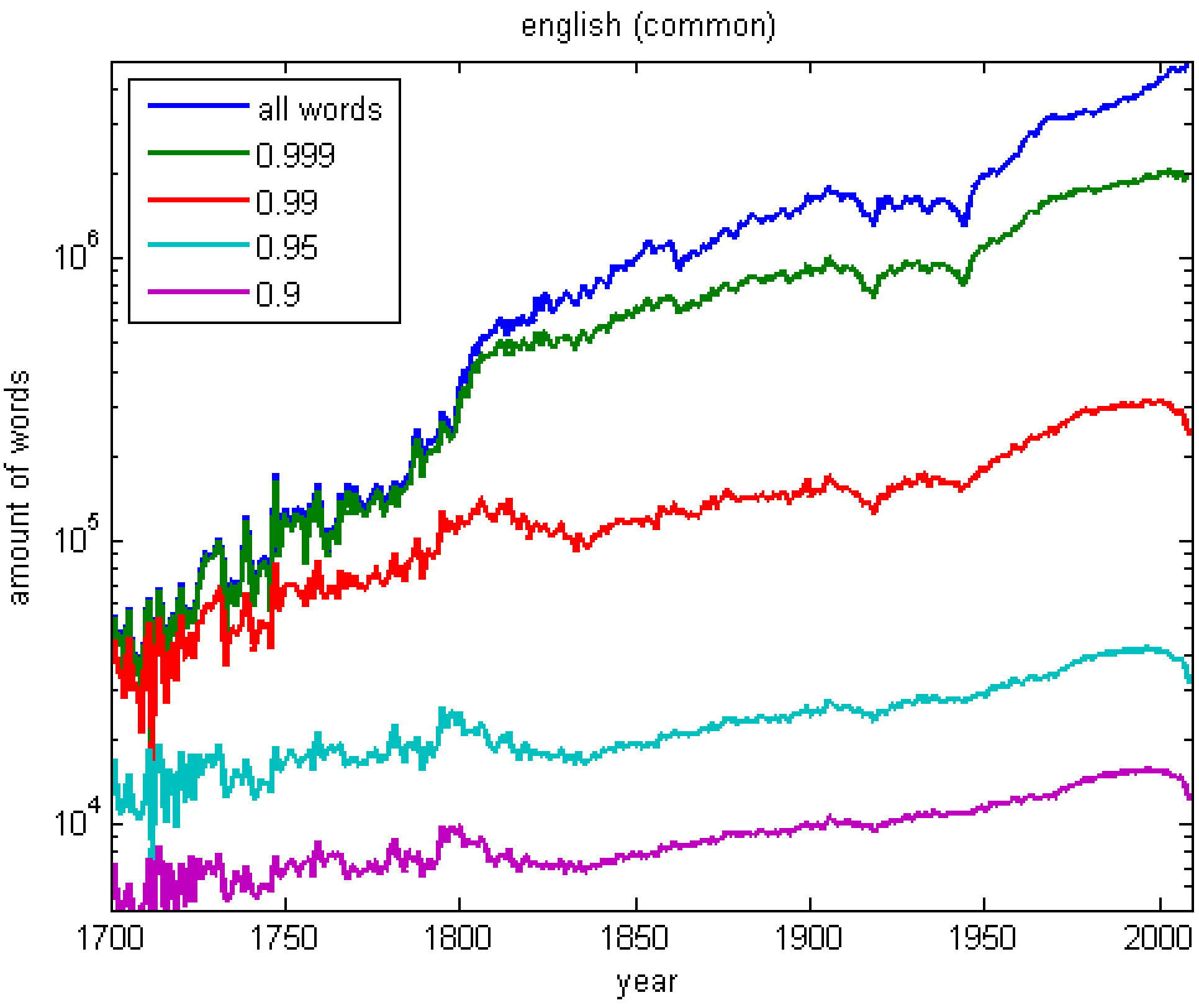

Fig. 4. The total amount of words used in different periods (the blue line). The share of words which contributed mostly to the selected word group is also shown

It can be clearly seen that rate of vocabulary growth is near constant within at least two centuries and makes about 400 words a year. That’s why this factor can’t explain why word length increases rapidly in the 20th century, stays constant in the 19th century and  decreases in the 21th century, although new words tend to appear.

It`s interesting to compare our vocabulary data with the data presented in the paper by Michel et al. It shows rapid growth of neologisms in the second half of the 20th century – on average 8500 words a year which results in the fact the number of English words increased for 1 million. Such difference can be explained by the fact that all the words with frequency of more than one billion, in other words very rare ones, such as scientific terms, were examined in the paper by Michel et al.

As hypothesis, we suggest the following mechanism of social factors influence on average word length.

Average word length which grows systematically over a large period of time reflects active process of word (term) creation which is connected with a certain idea which dominates in society. Actuality of basic principles of the new idea, which are verbalized by a particular set of words, leads to a significant growth of their usage. These terms are usually presented by multisyllable words. It`s approved by the fact that new concepts are complicated and new words which define them are formed with the help of affixation, for example *socialism, globalisation.* Besides, as it has already been said, the number short words is limited in any language and they can't be used to verbalize all the notions. All this leads to average word length growth.

Decrease in usage of notional words is explained by the loss of actuality of a certain lexical field which, in its turn, determines average word length on the whole. Language economy principle should also be mentioned which also contributes to decrease in word average length. According to this principle, any language tries to transmit more information using fewer linguistic means. As for terminology, the following situation can be seen. Phonetic and correspondingly orthographical structure of a term during the process of its assimilation to public needs is reduced due to the processes of abbreviation, contraction, substitution of long set expressions by shorter terms.

There is no prevalent idea during the time where average word length is relatively constant, and change of average word length is sporadic.

In other words, change of dominating concept results in rapid change of the word stock connected with it. Graphically it is expected to show change in curve incline angle, or appearance of small 'gaps' which reflects the situation when old concept is not topical any more and the new one has`t formed yet. If average word length grows during significantly long period of time according to the law close to linear (for example, in English from 1900 to 2000), it can be expected that rapid social concept change isn't taking place at that time. It slowly evolves over this period of time.

**Hypothesis verification. Evolution of notional words**

Let`s start the survey from examining notional words over the period of 1990-2000. Below, there is a list of words which contributed mostly to average word length increase during this period of time. The number shows word contribution level to total average length increase. The contribution is calculated according to the formula (5).

1. development 0.00648
2. information 0.00632
3. international 0.00596
4. university 0.00521
5. political 0.00415
6. relationship 0.00413
7. economic 0.00371
8. research 0.00346
9. production 0.00310
10. significant 0.00304

It can be seen that semantically these words are close to the key idea of the 20th century and reflect priorities of the English speaking society of that time.

Let`s take the following approach to verify the above hypothesis. Let`s divide the interval from 1800 to 2000 into some smaller ones of 25 years each (an average age of a certain generation), the last interval from 2001 to 2008 contains the data of only the stated period. (Hopefully, Google Books will be enlarged in future and new data will be available).

Each word contribution to change of average word length is calculated at each interval. The contribution is calculated according to the formula (5).

Appendix 1 show notional words which contributed mostly to average word length increase or decrease for every these periods.

Some remarks should be made: 1. There are a lot of mistakes at the 1st interval (1800-1825) caused by wrong symbol recognition account for low quality of print and poor condition of books. In particular, letters *s* and *t* are often recognized as *f*. It introduces some errors but isn`t of great relevance for overall conclusion. 2. The number of short notional words is quite small, but it doesn`t contribute much to the common value. Later, they won`t be regarded. 3. Words beginning both with capital and small letters are regarded, no matter that the tables contains words beginning with capital letters.

For further analysis, the most "influential" words are distinguished. These are the words which belong to the first ten words contributed greatly to word average length decrease, or to the first ten words contributed mostly to word average length decrease at least during two periods.

The data are summed up in tables 1, 2. '+' sign means that the word belongs to the first ten "influential" words.

Table 1. Notional words which contributed mostly to average word length

| Word | 1800-1825 | 1825-1850 | 1850-1875 | 1875-1900 | 1900-1925 | 1925-1950 | 1950-1975 | 1975-2000 | 2000-2008 |
|---|---|---|---|---|---|---|---|---|---|
| One | | + | + | | | | | | |
| Position | | + | + | | | | | | |
| Government | | + | + | | | + | | | |
| Development | | + | + | + | + | + | + | | |
| American | | + | | + | | + | | | |
| Conditions | | | + | + | + | | | | |
| International | | | | | + | + | | + | |
| Production | | | | | + | + | | | |
| Individual | | | | | + | + | | | |
| Education | | | | | + | | + | | |
| Political | | | | | | + | + | | |
| University | | | | | | | + | + | |
| Information | | | | | | | + | + | |

Let`s analyse these data. Special attention should be paid to the fact that word lists at the beginning of the 19 and 21 centuries are not similar to word lists of other periods. Language development in 2000-2008 will be regarded later. As for the beginning of the 19 century, obviously, some interesting and important events took place at the turn of the century but 18 century data are needed to survey them. Unfortunately Google Books doesn`t contain enough data. This question isn`t regarded in this paper.

The word ‘development’ is at the top of word list which contributed to average word length over 1.5 centuries. This shows a key role of this concept for English speaking society.

Besides ‘development’, the words which are common for different periods of the 19 centuries are *one, position, government, American, conditions*. They belong to different semantic groups and it isn`t clear to what single idea they may belong to.

As for the words from table 1, which are common for different periods of the 20 century, except ‘development’, they are the following: *international, production, individual, education, political, university, information*. It seems that these words are key words in the modern English-speaking world which characterize the global economical and political system based on education and information. There are some common ‘+’ marked words among different periods of the 20th century presented in table 1, which indicates that the system evolves gradually.

Absolute values of contribution to average word length also attract attention. As for the 19$^{th}$ century, the contribution value is 0.003 (except 1800-1825), as for the 20th century the contribution value is 0.005 (except 2000-2008). In other words, the frequency of *international, production, individual, education, political, University, information* decreased in the 20$^{th}$ century more rapidly than *of one, position, government, American, conditions* in the 19$^{th}$ century. It shows that correspondent concepts were more significant for society in the 20th century than in the 19th.

It`s interesting that the words which contribute most to average word length increase in 21 century don`t belong to any concept.

Thus, the analysis of notional words contributed mostly to word frequency increase supports this hypothesis.

Let`s consider the words which frequency decreased rapidly. The data are shown in table 2.

Table 2. Notional words which contributed to average word length decrease

| Word | 1800-1825 | 1825-1850 | 1850-1875 | 1875-1900 | 1900-1925 | 1925-1950 | 1950-1975 | 1975-2000 | 2000-2008 |
|---|---|---|---|---|---|---|---|---|---|
| Particularly | | + | + | | | | | | |
| Immediately | | + | + | + | | | | | |
| God | | + | + | + | | | | | |
| Circumstances | | | + | + | + | | | | |
| Character | | | + | + | + | + | + | + | |
| Christian | | | + | | + | | | | |
| Practically | | | + | | | + | + | | |
| Mr | | | | + | + | + | | | |
| Hundred | | | | | + | + | | | |
| Constitution | | | | | + | | + | | |
| Little | | | | | | + | + | | |
| Necessary | | | | | | | + | + | |

The data for the beginning of the 19th and 21st centuries stand apart. It`s interesting that frequency distribution of word *character* constantly decreases. On the whole, the words presented in this table belong to different semantic fields and don`t reflect any particular idea.

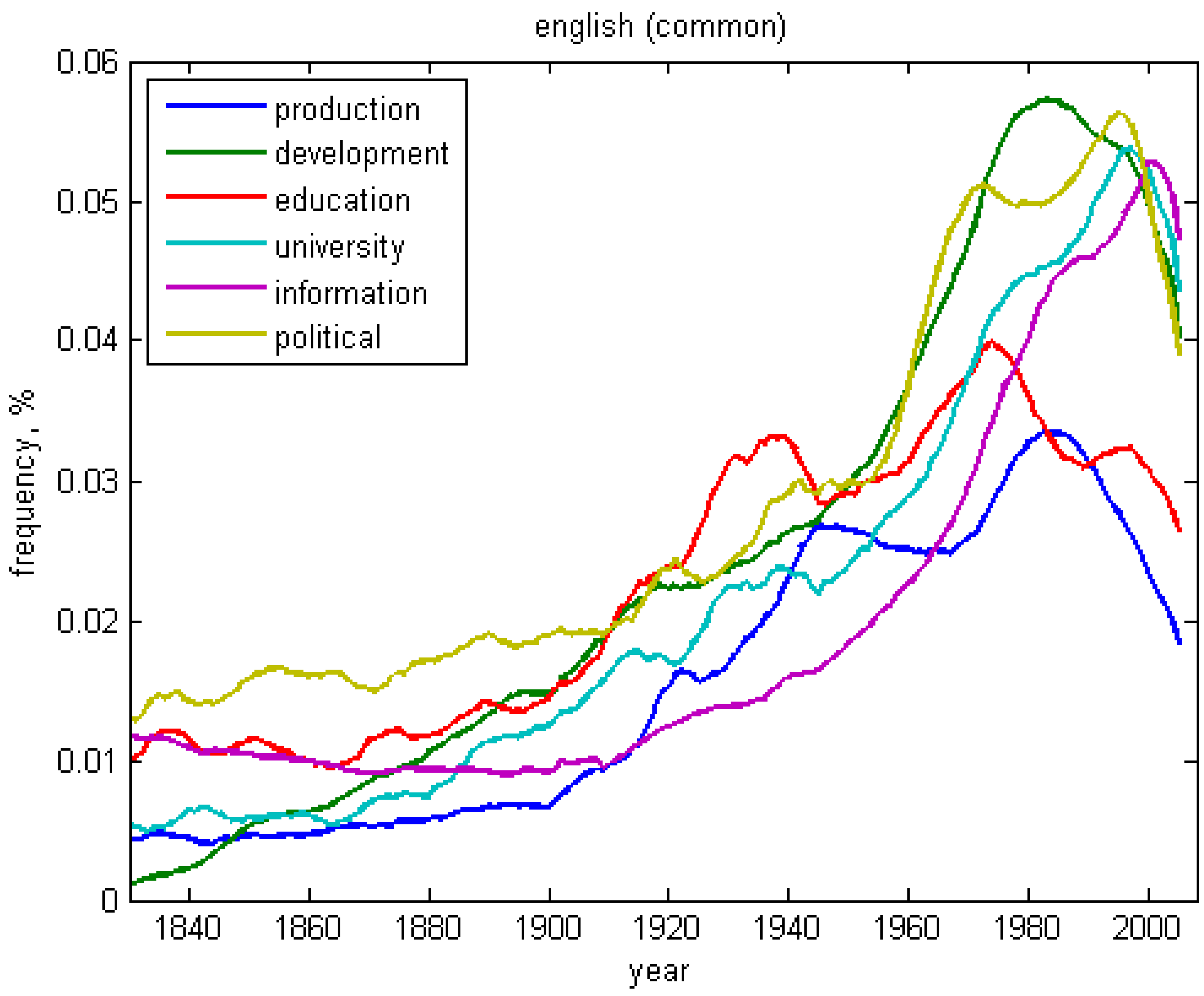

Figure 5. Dynamics of the top ten influential words of the $20^{th}$ century

Absolute values of contribution to average word length of words also attract attention. Both for the 19th and 20th century, the contribution value is 0.0025 (except 1800-1825). This value is much bigger over 1800-1825, 2000-2008 and approximately is 0.01. Words which frequency decreases greatly in the 21st century are of special interest. The less-frequently used words are the following development, university, international, political, information which frequency increased rapidly during the previous years according to table 1. That`s why the dominant idea of the 20th century doesn`t prevail any more. Thus, the analysis of notional words contributed less to word frequency supports this hypothesis.

Let`s consider the diagrams of the top ten influential words dynamics in the 20th century: development, information, international, University, political, relationship, economic, research, production, significant.

It is clearly seen that these diagrams reflect the common rules of average word length change in the 19-21 centuries. During the 19th century, the majority of these words (except development) don`t increase in frequency and even decrease in it. Their frequency increases in the 20th century. In the 21st century the frequency of these words decreases, moreover decrease in frequency of some words (development, production) began earlier, approximately in 1980.

**Evolution of function words**

Let`s consider functional words. They form a set of words which contains several dozens of words. All the functional words which we distinguish (which can be found in appendix 1) are the following: articles, prepositions, conjunctions, pronouns, auxiliary verbs, particles and some proverbs.

Below, there is a list of words which contributed mostly to word length increase during the 20th century.

1. the 0.04110
2. of 0.03947
3. he 0.02033
4. it 0.01848
5. I 0.01688
6. his 0.01307
7. to 0.01136
8. was 0.00932
9. at 0.00923
10. and 0.00886

This list of functional words consists mostly of personal pronouns (4 words). The dynamics of personal pronouns is rather interesting.

It can be clearly seen that frequency of the majority of these words is stable or decreases rapidly till the end of the 20th century, after which frequency of all the pronouns increases rapidly. As all of them are short, decrease in their frequency contributes to average word length increase and increase in their frequency in the beginning of 21 century contributes to average word length decrease. That`s why this factor has the same influence as the evolution of notional words which was described earlier. But it`s not clear whether these two factors are mutually connected.

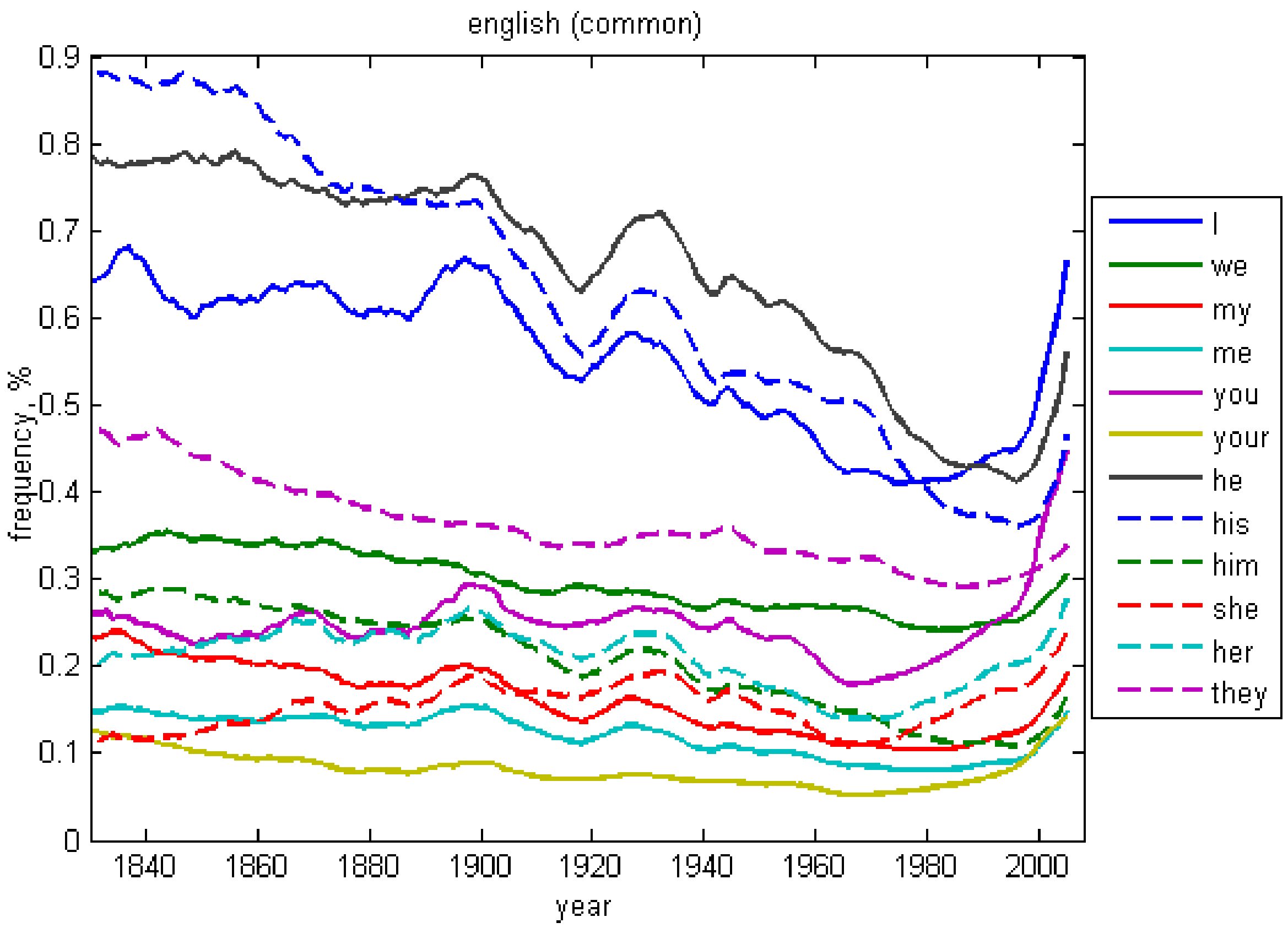

Figure 6. Dynamics of personal pronouns in English

The majority of other functional words are used with the frequency near constant. Some of them show the same rules of frequency change as personal pronouns but this tendency is much weaker.

**Average word length dynamics in Russian**

Let`s consider Russian words to text our conclusions. It`s interesting to compare English with the Russian because Russian society differs greatly from English-speaking society and had completely different views and values during almost a century. Average word length dynamics in Russian is shown in figure 7.

As it can be seen, the situation is almost the same as in the English language: it is near constant in the 19th century, grows in the 20th century, but the decrease starts earlier - about in 1975.

The Russian language data in the same format as the English ones are given in appendix 2. Let`s consider the most interesting period – the 20$^{th}$ and the beginning of the 21$^{th}$ century. It should be noted that the chosen time periods for both languages (1900-1925 etc.) are not satisfactory for Russian because such important events in Russian history as the socialist revolution (1917), the World War II and the breakup of the Soviet Union took place in the middle of the intervals which led to some vocabulary mixing during the corresponding periods. Anyway, the results for Russian are rather clear.

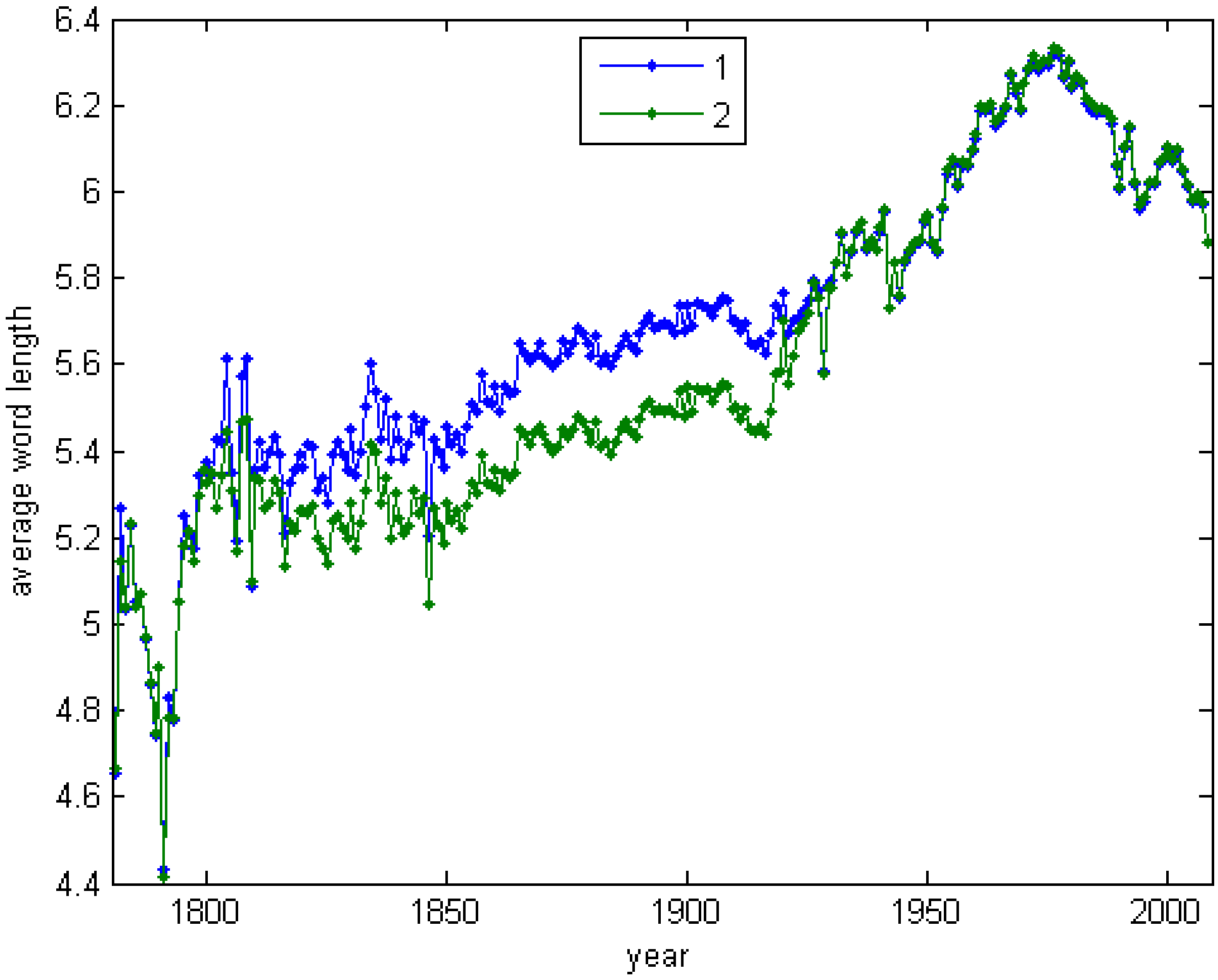

Fig. 7. The blue curve shows average word length taking into consideration the orthographical rules of that time, the green one shows it excluding the letter «ъ» at the end of each word (rule of 1917 year)

In distinction from English, the top ten influential words for the chosen periods which contributed to average word length almost don`t coincide. Different concepts are presented at each interval. Over the period 1900-1925 economic problems which dates back to the 19$^{th}$ century mixes with the revolutionary ideas. During 1925-1950 the most influential words are articulated on ideas of socialist regime and Soviet power and key for country problems of that time concerning construction and agriculture. The priorities changed in 1950-1975, they concern manufacture and management. The leading concept was changed again in 1975-2000, it articulates on the idea of nationhood. The idea of legality dominates in 2000-2008.

It`s typical that during 1975-2000, the words which were connected with manufacturing and socialism were the words which contributed greatly to average word length in previously periods during this period. Concept change between the intervals 1925-1950 and 1950-1975 can be seen as a small gap on the diagram a bit earlier than 1950.

Figure 8 shows diagrams for the top ten words. In distinction from English Russian lines in Fig.8 are not so smooth as in English. Many of them were used widely only after the revolution. Increase in frequency was broken by the World War II where the USSR was involved in 1941. It should be noted that priorities were changed during the 50-s: problems of effectiveness became acute and the term *revolution* fell by the wayside. It`s interesting that decrease in frequency of influential Russian words started in 1980 as well as in English.

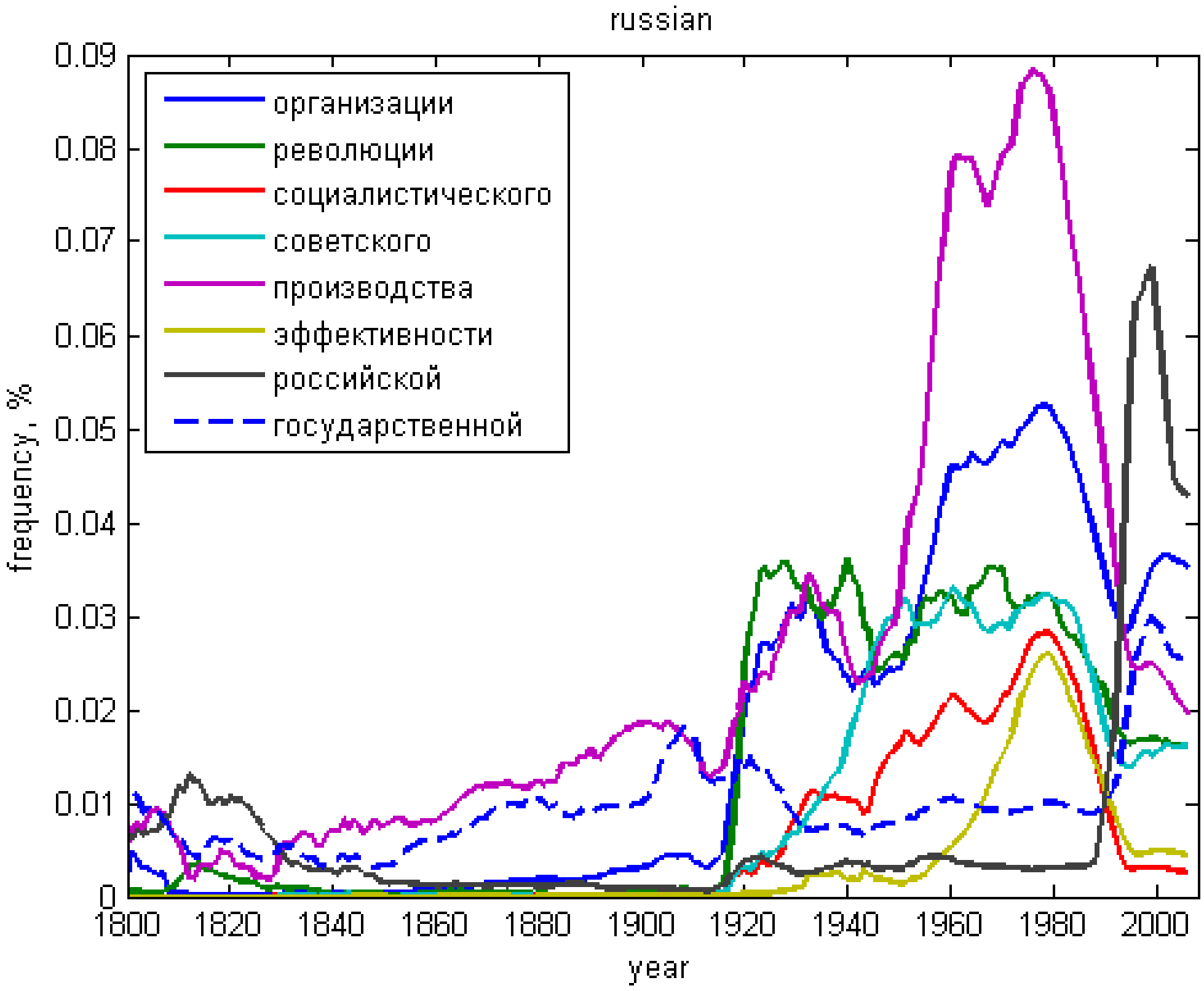

Figure 8. Dynamics of the top ten influential Russian words of the 20th century

Thus, there is direct dependence between fluctuations of average word length and distinguishing of word clusters which verbalize the dominating concept.

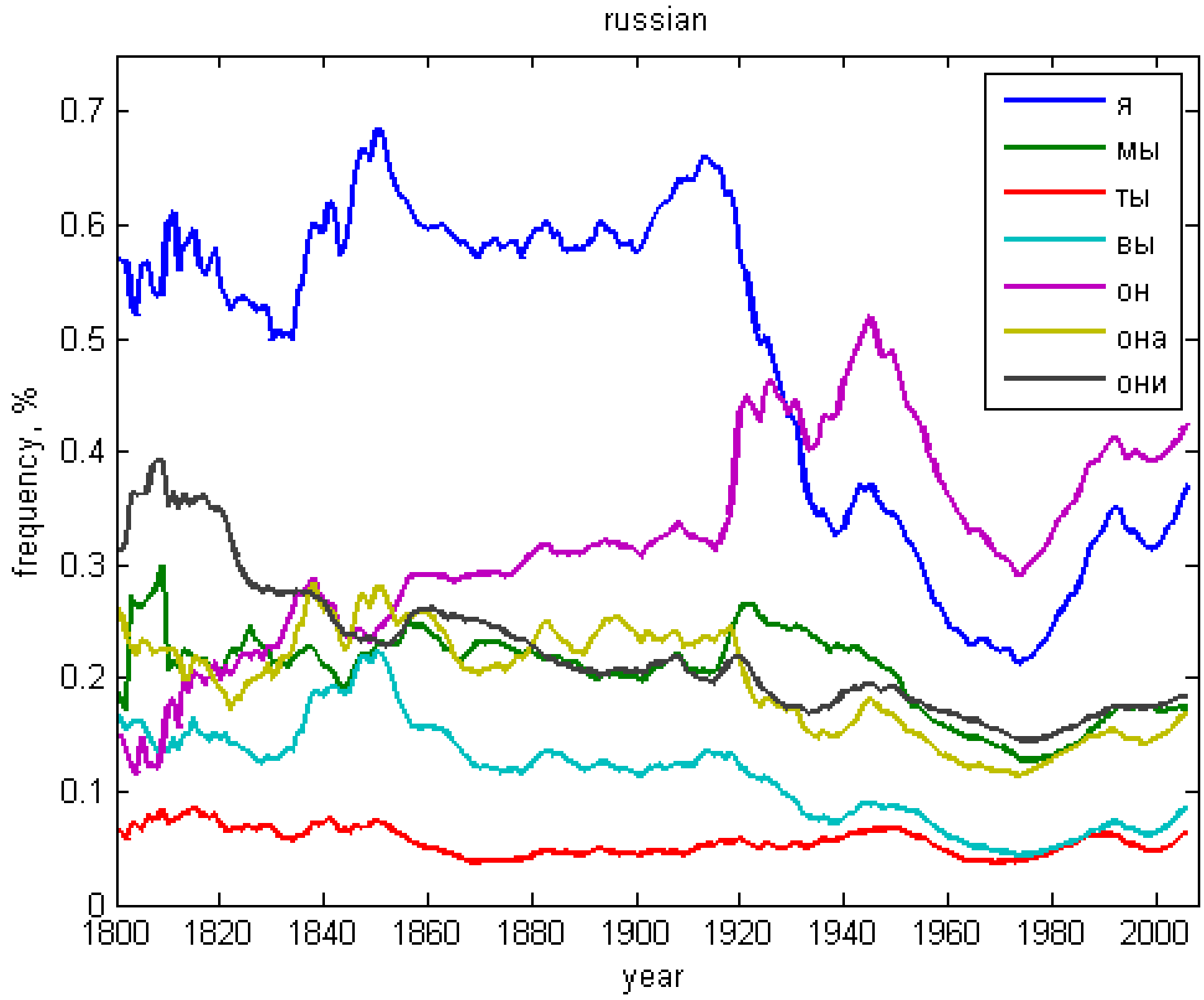

Figure 9. Dynamics of personal pronouns in Russian

As for functional words, Russian personal pronouns behave as English personal pronouns (figure 6) and decrease in frequency takes place in 1975.

Rapid decrease in frequency of the personal pronoun *я* (*I*) in the Russian language after 1917 and till 1975 can be explained by detraction of the role of personality in the time of socialism and imposture of collectivism ideology. Again it is not clear why grammatical changes occur simultaneously with lexical ones.

**Conclusion**

The paper introduces data concerning change of average word length in the English and Russian languages during the last two centuries based on e-library Google Books and system Ngram Viewer. It is shown that this parameter is a result of change in word frequency. Change in frequency of content and functional words is analysed separately.

It is shown that decrease in average word length correlates with two social-cultural factors: existence of some fundamental idea or several ideas which is connected with decrease in personal pronouns.

Decrease in average word length correlates with retreat from ideas which has been dominating (the number of words which verbalize this concept decreases rapidly), and the frequency of personal pronouns increases. When there are no prevailing concepts, word average length stays almost the same.

It`s interesting that the obtained data correlate well both in English and Russian in spite of great social, cultural, economical and political differences between their speakers. It shows that the traced regularities are not accidental but , probably, cognitive.

Ngram Viewer system enables studying of thin language evolution using change of frequency data as a basic mechanism reflecting the influence of pragmatic factors.

This work was funded by the Russian Foundation for Basic Research through project RFBR № 12-06-00404-a.